\documentclass{article}
\usepackage{spconf,amsmath,graphicx}
\usepackage{booktabs}
\usepackage{url}
\usepackage[compress]{cite}
\usepackage{xspace}
\newcommand*{\eg}{e.g.\@\xspace}
\newcommand*{\ie}{i.e.\@\xspace}
\usepackage{xcolor}
\usepackage{soul}

\newcommand{\comment}[1]{}
\usepackage{multirow}
\usepackage[hidelinks]{hyperref}

\newcommand{\GRIDPictureExamples}[1]{%
\begin{figure}[#1]
\vskip 0.2in
\begin{center}
\centerline{\includegraphics[width=0.4\columnwidth]{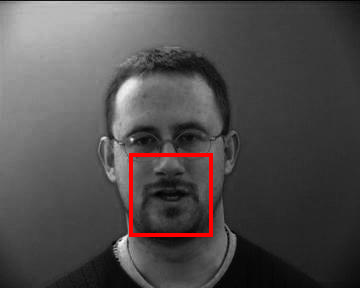}\includegraphics[width=0.4\columnwidth]{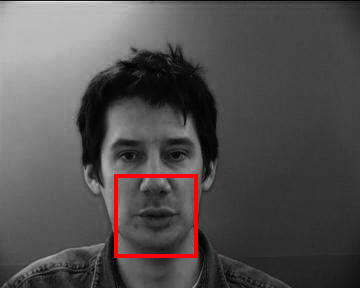}}
\caption{Two randomly chosen example frames from the GRID corpus with highlighted mouth area.}
\label{fig:GRIDExample}
\end{center}
\vskip -0.2in
\end{figure}}

\newcommand{\ResultsAvg}[1]{%
\begin{table}[#1]
\caption{Word Accuracy averaged over the Development Speakers 1--9 and Evaluation Speakers 10--19.}
\label{tab:results}
\begin{center}
\begin{tabular}{cccc}
\toprule
\textbf{Spk.}&\textbf{Method} & \textbf{Word Acc.} & \textbf{Std. Dev.} \\
\midrule
\multirow{3}{*}{\rotatebox[origin=c]{90}{1--9}}&Eigenlips + SVM      & 68.4\% & 7.4\% \\
&HOG + SVM            & 71.1\% & 6.7\%  \\ 
\cmidrule{2-4}
&LSTM   & 79.4\% & 4.3\% \\
\midrule
\multirow{3}{*}{\rotatebox[origin=c]{90}{10--19}}&Eigenlips + SVM      & 70.6\% & 4.2\% \\
&HOG + SVM            & 71.3\% & 4.3\%  \\ 
\cmidrule{2-4}
&LSTM   & 79.6\% & 4.3\% \\
\bottomrule
\end{tabular}           
\end{center}
\end{table}
}

\newcommand{\FigureConfMx}[1]{%
\begin{figure}[#1]
\vskip 0.2in
\begin{center}
\includegraphics[width=\columnwidth]{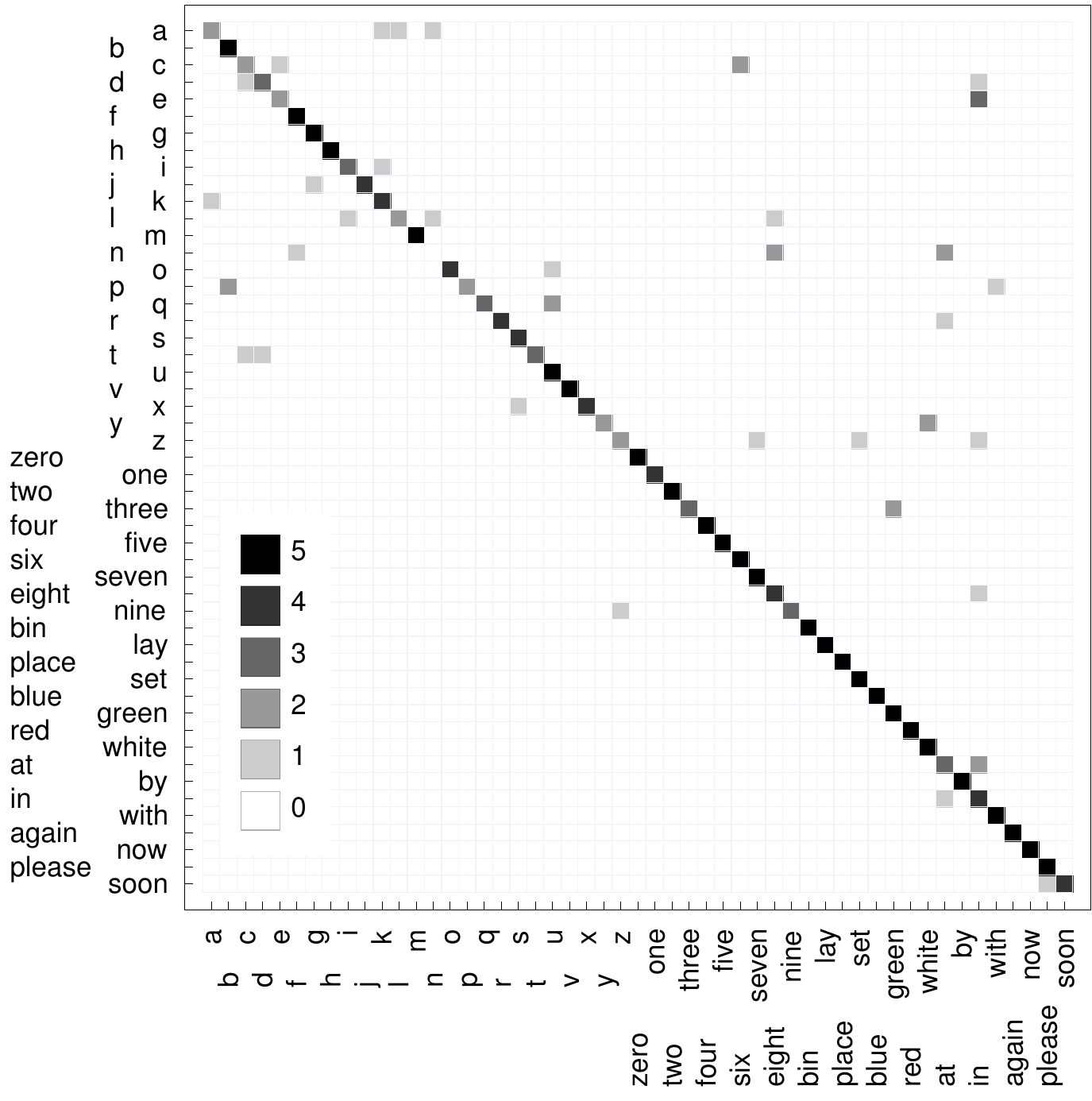}
\caption{Confusion matrix for speaker 7, using the neural network lipreader. Rows represent reference labels, columns represent the
classification hypothesis. Greyscale level indicates the density of matching reference and hypothesis.
The accuracy is highest on longer words and lowest on letters.}
\label{fig:ConfMx}
\end{center}
\end{figure}}

\title{Lipreading with Long Short-Term Memory}
\name{Michael Wand, Jan Koutn\'{i}k, J\"{u}rgen Schmidhuber\thanks{This research was supported by the FP7 Marie Curie Initial Training Network PROTOTOUCH (grant \#317100) and the Swiss National Science Foundation grant ``Advanced Reinforcement Learning'' (grant \#156682).}}
\address{The Swiss AI Lab IDSIA, USI \& SUPSI}
\begin{document}

\maketitle
\begin{abstract}
\textit{Lipreading}, \ie speech recognition from visual-only recordings of a speaker's face, can be achieved with a processing pipeline based solely on neural networks, yielding significantly better accuracy than conventional methods.
Feed-forward and recurrent neural network layers (namely Long Short-Term Memory; LSTM) 
are stacked to form a single structure which is 
trained by back-propagating error gradients through all the layers. 
The performance of such a stacked network was experimentally evaluated and compared to a standard Support Vector Machine classifier using conventional computer vision features (Eigenlips and Histograms of Oriented Gradients). 
The evaluation was performed on data from 19 speakers of the publicly available GRID corpus. 
With 51 different words to classify, we report a best word accuracy on held-out evaluation speakers of 79.6\% using 
the end-to-end neural network-based solution (11.6\% improvement over the best feature-based solution evaluated). 
\end{abstract}
\section{Introduction}
\label{sec:introduction}
It is well-known that humans understand speech not only by listening, but also by
taking visual cues into account \cite{McGurk_HearingLips}. Hearing-impaired persons are in fact able 
to comprehend human speech by purely visual \textit{lipreading}, \ie by processing visual information from a speaker's lips and face.
Consequently, research on making lipreading available to
electronic speech recognition and processing systems has been of interest for some decades, with pioneering work done by Petajan
\cite{Petajan_PhD_Lipreading}. 
His PhD thesis proposed to use lipreading
to augment conventional automatic speech recognition (ASR), yet later researchers started to 
perform \textit{purely visual} speech recognition \cite{Chiou_LipreadingFromColorVideo}, which is 
also the goal of this study. Lipreading systems typically 
consist of (at least) \textit{feature extraction} and \textit{classification}.
The feature extraction can become quite complex: Many recent lipreading systems,
\eg \cite{Lan_AVSP09_VisualFeaturesLipreading,Bowden_SPIE13_RecentDevelopmentsLipreading},
use a lip tracking system as a first stage, followed by versatile image features such as Active Appearance Models \cite{Cootes_ActiveAppearanceModels}
or Local Binary Patterns \cite{Zhao_LipreadingSpatiotemporalDescriptors}.
Classification is frequently done with Support Vector Machines (SVMs), \eg \cite{Zhao_LipreadingSpatiotemporalDescriptors},
or Hidden Markov Models (HMMs), \eg 
\cite{Hueber_IS07_UltrasoundOpticalContinuousPhoneReco,Bowden_SPIE13_RecentDevelopmentsLipreading,Busso_IS14_LipreadingIsolatedDigits,Lan_AVSP09_VisualFeaturesLipreading}.

Our aim is to replace the complete visual speech recognition pipeline with a compact neural network architecture. Neural networks (NNs) have become
increasingly popular in conventional speech recognition, first as feature extractors in an HMM-based 
architecture \cite{Bourlard_ConnectionistSpeechReco,Hinton_DNNsForAcousticModeling,Graves_ASRU13_HybridASRDeepBLSTM}, more recently 
replacing the entire processing chain
\cite{Graves_ICASSP13_SpeechRecognitionDeepRNN}. For the latter, the 
\textit{Long Short Term Memory} (LSTM; 
\cite{Hochreiter_LongShortTermMemory})
architecture is typically used. 
Consequently, our approach to the lipreading problem uses a NN that chains feed-forward layers and LSTM layers, described in detail 
in \autoref{ss:networks}. Manual feature extraction is no longer required. The NN inputs are now the raw mouth images, as is 
common in modern computer vision tasks, but stands in stark contrast 
\eg to \cite{Zhao_LipreadingSpatiotemporalDescriptors,Bowden_SPIE13_RecentDevelopmentsLipreading}.

\section{Related Work}
\label{sec:related}

Lipreading has been used 
as a complementary modality for speech recognition from noisy audio data \cite{Petajan_PhD_Lipreading,Bregler_ICASSP94_Eigenlips},
as well as for purely visual speech recognition \cite{Chiou_LipreadingFromColorVideo,Bowden_SPIE12_VideoToText,Noda_IS14_LipreadingUsingConvNet}. The latter gives rise to a 
\textit{Silent Speech interface}, which is defined as a system ``enabling speech communication to take place when an audible acoustic signal is unavailable'' \cite{Denby2010}.
Silent Speech technology
has a large number of applications: It allows persons with certain
speech impairments (\eg laryngectomees, whose voice box ({\em larynx}) has been removed) to  communicate, as well as enabling confidential and undisturbing communication
in public places \cite{Denby2010}. Further uses of lipreading have been proposed, \eg automatic speech 
extraction from surveillance videos and its interpretation for forensic purposes \cite{Bowden_SPIE13_RecentDevelopmentsLipreading}.
Lipreading has been augmented with \textit{ultrasound} images of the tongue and vocal 
tract \cite{Denby_ICASSP04_UltrasoundSpeechSynthesis,Hueber_ICASSP07_EigentongueUltrasound,Hueber_SSIUltrasound}.
Furthermore, there are Silent Speech interfaces based on very different principles, like speech recognition from electromyography
\cite{Sugie_SpeechProsthesis_85,Morse_EMGToRecognizeSpeech_89,SchultzWand_SCJ10_Coarticulation,Wand_EMGSpeakingModeVarieties}
or (electro-)magnetic articulography \cite{Fagan_SilentSSR}.

NNs have been used in speech recognition as feature extractors in HMM-based speech recognizers 
\cite{Bourlard_ConnectionistSpeechReco,Hinton_DNNsForAcousticModeling}. Neural networks, LSTMs in particular, 
started to replace larger parts of the speech processing chain previously dominated by HMMs. 
An end-to-end neural network system~\cite{Graves_ICML06_CTC,Graves_ICML14_EndToEndNeuralNetworkSpeechRecognition} finally outperformed 
HMM-based systems and achieved the best performance (16\% error) on the large Switchboard Hub5'00 speech recognition benchmark~\cite{Hannun_DeepSpeech}.

Massively parallel graphics processing units (GPUs) became available in the last few years. Since then, {\em Convolutional} NNs (CNNs) trained by 
gradient descent~\cite{Lecun_BackpropagationForZipRecognition} dominate, 
e.g.~\cite{Ciresan_IJCNN11_TrafficSignClassification}, the area of 
image recognition, 
as well as related tasks like object detection and segmentation. The first CNN 
application in lipreading \cite{Noda_IS14_LipreadingUsingConvNet} uses the CNN as a pre-processor for an HMM-based sequence classifier.

\section{The GRID Data Corpus}
\label{sec:corpus}

\GRIDPictureExamples{t}

Our experiments were performed using the GRID audiovisual corpus \cite{Cooke_GRIDCorpus}\footnote{Publicly available at \url{http://spandh.dcs.shef.ac.uk/gridcorpus}}, 
consisting of video and audio recordings of $34$ speakers saying $1000$ sentences each. 
The total length of the 
recordings is 28 hours; two example video frames are shown in \autoref{fig:GRIDExample}. Each of the 
sentences has a fixed structure: \textit{command(4) + color(4) + preposition(4) + letter(25) + digit(10) + adverb(4)}, for example ``Place red at J 2, please'', where
the number of alternatives words is given in parentheses. 
A total of $51$ different words are contained in the GRID corpus;
the alternative words for each of the six sentence parts are distributed uniformly. 
The letter \textit{W} is excluded because its pronunciation is vastly longer than for any other letter.

Sentences have a fixed length of 3 seconds at a frame rate of 25 frames per second. Each sentences thus spans across 75 frames.  Video data
bis available in ``normal'' and ``high'' quality; the normal quality video with $360\times 288$ pixel resolution, converted to greyscale, was used.
Unreadable videos, in particular for speaker number 8, were discarded.
The frame-level alignments distributed with the corpus were used to obtain word level segmentations of the video,
causing the training dataset to consist of $6 \cdot 1000 = 6000$ single words per speaker.
The acoustic part of the GRID corpus was not used. 

A $40\times 40$ pixel window containing the mouth area from each video frame was extracted using the following procedure. The face area localized using the Mathematica {\tt FindFaces$[\,\,]$} function was converted into the LAB color-space. The {\it A} component pixels were multiplied element-wise by a Gaussian matrix (with mean located at 30\% of of the image height along the middle column) and $\sigma = 500$ pixels, and rescaled into $[0,1]$ interval. The center of mass of pixels that have value above $0.9$ was considered as the mouth center. The face area size was inflated by factor of 1.5, scaled to 128 pixel width, and a window of $40\times 40$ pixels with the mouth coordinates in the center was extracted. This patch was converted to greyscale, the contrast was maximized (i.e. all pixel values were remapped to $[0,1]$ interval), and all the values in the complete dataset were standardized.

Speakers 1--19 from the entire GRID were used: speakers 1-9 form the development set that was used to determine the parameter settings;  
speakers 10--19 form the evaluation set, held back until the final evaluation of the systems.
All experiments are \textit{speaker-dependent}, \ie training and test data for the
classifiers were always taken from the same speaker. The results reported in this paper are averaged over the speakers. The data for each speaker was randomly divided into training, validation,
and test sets, where the latter two contain five sample videos of each word, \ie a total of $51 \cdot 5 = 255$ samples each.
The training data is however highly unbalanced:
For example, each letter from ``a'' to ``z'' appears 30 times, whereas each color appears 240 times.

\section{Methods}
\label{sec:methods}

\subsection{Baseline Feature Extraction and Classification}
\label{ss:baseline}

The NN-based lipreader was compared to a baseline SVM classifier using conventional features, namely 
\textit{Eigenlips}~\cite{Bregler_ICASSP94_Eigenlips}, which were used as a baseline feature in \cite{Noda_IS14_LipreadingUsingConvNet}, and
\textit{Histograms of Oriented Gradients} (HOG) \cite{Dalal_CVPR05_HistogramsOrientedGradients} as
a more complex feature which  yielded good performance in preliminary experiments.

Eigenlip features are created from raw frames by computing the PCA decomposition on the training data and then transforming all the images by multiplication with the PCA matrix, retaining only
 a certain number of dimensions ordered by maximal variance.
HOG is originally a feature extractor for 
object recognition \cite{Dalal_CVPR05_HistogramsOrientedGradients};
it divides the image window into small spatial
regions (cells) and accumulates a local 1-D histogram 
of gradient directions or edge orientations over the
pixels in each cell. Histogram entries are normalized over larger spatial areas. The HOG features were obtained using the VLFeat library \cite{Vedaldi_VLFeat}.

Since SVMs are not sequence classifiers, a single feature vector has to be computed from the sequence of frames representing each word,
called \textit{sequence feature vector}.
Sequences vary in length. For example, a typical letter ``a'' is pronounced in 3--4 frames, whereas a longer word like
``please'' can occupy more than 10 frames. In the sequence feature vector, all frames are stacked while enforcing a
specified vector length: frames are repeated if a sequence is shorter than this length, neighboring frames of longer sequences are averaged.

\subsection{Neural Network Lipreader}
\label{ss:networks}

Neural networks (NNs) consist of processing units (neurons) connected by trainable weighted connections. The neurons are typically 
organized in layers, which can be broadly distinguished as follows based on 
their connectivity: (1) feed-forward NNs pass the input signal to 
the output neurons without allowing cyclic computations; (2) recurrent NNs wire the connections in a cycle which forms a temporal memory. NNs are, in the supervised case, typically trained by gradient descent, which is realized by error back-propagation through the layers followed by adjustment of the weights. In the case of recurrent NNs the error propagates along the time axis as well, referred to as back-propagation through time, implemented by unfolding the recurrent connections into a feed-forward structure as deep as the length of the sequence. 
Such deep networks cause the gradient to explode or vanish~\cite{Hochreiter_GradientFlow}, which can be fixed by replacing a single recurrent NN unit by a LSTM cell that avoids the problem by linear recurrent connection of a {\it cell} inside the LSTM unit \cite{Hochreiter_LongShortTermMemory}. The information flow through the LSTM cell is regulated by \textit{input}, \textit{output} and \textit{forget} \cite{Gers:2000nc} gates using multiplicative connections. See~\cite{DBLP:journals/corr/GreffSKSS15} for detailed LSTM description, analysis and setup.

\section{Experiments}
\label{sec:experiments}

\subsection{Experimental Setup}

Two feature extractions (Eigenlips and HOG), both combined with the SVM classifier,
were compared to the LSTM lipreader. All parameters were optimized
on the development speakers 1--9. The error on all systems is reported on the speaker-dependent test set, no training error is reported.

\ResultsAvg{t}

In a series of experiments on the speakers 1--9, the optimal configuration of the 
feature extraction for the SVM experiments was determined, as well as the best neural network structure.
The best PCA cutoff for the Eigenlip features was at 100 components, the best HOG cell size
was 8.
Best SVM recognition results were obtained with a linear kernel at a sequence 
feature vector length of 6 frames. Increasing the feature vector length 
did not improve the accuracy, almost certainly because
among the short 
video sequences containing letters, where most errors occur (see \autoref{ss:results}), 6~frames already 
cover the entire information available. 
Taking this observation into account, we hypothesize 
that a dedicated sequence classifier like an HMM would \textit{not} substantially 
improve the classification accuracy on this corpus.

The input data for the SVM
can become very high-dimensional, which we assume to be the reason why higher-degree polynomial 
SVM kernels yielded lower accuracy than linear kernels.
Zhao et al.~\cite{Zhao_LipreadingSpatiotemporalDescriptors} report that second-degree polynomial SVM kernels 
perform the best, however their SVM classifier treats the sequential information differently.

The best LSTM lipreader consists of one feed-forward layer followed by two recurrent LSTM layers,
128 units (neurons / LSTM cells) each,
and a softmax layer with 51 units that perform the word classification.
The learning rate was set to $0.02$, momentum was not used, and early stopping with a delay of 10 epochs was used.
Weights were initialized from uniform distribution over the range $[-0.05,0.05]$. 

\FigureConfMx{t}

\subsection{Results}
\label{ss:results}

The word level classification accuracies are summarized in \autoref{tab:results}. 
The LSTM lipreader yields statistically significant improvement (one-tailed t-test with $p=0.05$) over 
the conventional features combined with the SVM classifier.
The Eigenlip and HOG features perform similarly, both
worse than the LSTM lipreader. On the Evaluation Speakers 10--19, the LSTM lipreader improves
the accuracy by 11.6\%, compared to the best conventional solution (HOG + SVM).

\autoref{fig:ConfMx} shows a typical confusion matrix on speaker 7 data, where classification 
was performed with the LSTM lipreader. 
The rows show reference word labels, the columns show hypotheses. The confusion on letters is far higher (upper part of the matrix) than on longer words (lower part of the matrix).
For this speaker and configuration the accuracy on the letters is 69.8\% (at 4\% chance level), the accuracy on the non-letter words is 93.4\% (at 3.8\% chance level). The total
accuracy is 82.0\%. 

The discrepancy between letters and other words is caused by three major factors: First, the letters from 'a' to 'z' are  highly confusing even under optimal circumstances.
In particular, this applies to  voiced and voiceless versions of the same letter, like `p' and `b'.
Consequently, \textit{visemes} (visual units for recognition) frequently do not distinguish
between such similarly-looking sounds at all (see \eg \cite{Cappelletta_EUSIPCO11_VisemeDefintions} and the references therein).
Second, single letters video sequences are often are very short, sometimes consisting of only 3-4 frames. This means that very little data is available and also that the letter pronunciation is highly influenced by adjacent sounds. 
This context does \textit{not} help distinguishing different letters, since the parts of the GRID sentences are statistically independent from each other.
We note that \cite{Noda_IS14_LipreadingUsingConvNet} report phone accuracy on a corpus which consists of
\textit{whole words}: here the context plays a great role in \textit{improving} recognition.

The confusion matrices are qualitatively similar across all speakers and all experimental setups -- the longer words are recognized with close to 100\% accuracy, whereas confusion is highest on the letters.

\section{Conclusion}
\label{sec:conc}

This study shows that the neural network based lipreading system applied to raw images of the mouth regions achieves significantly better word accuracy than a system based on a conventional processing pipeline utilizing feature extraction and classification. 
The LSTM lipreader with a single feed-forward network, which learns the features automatically together with training the LSTM sequence classifier, consistently achieved almost 80\% word accuracy in speaker-dependent lipreading.

The experiments, not described in this paper, also included the highly popular CNNs instead of the fully connected feed-forward layer, but the results did not improve. One of the possible reasons is that the small, $40\times 40$ pixel area already  contains just enough information for the classification. Experiments with CNNs and large image sizes as well as evaluation of a speaker-independent LSTM lipreader are the subject of future experiments.

\bibliographystyle{IEEEtran} 
\bibliography{IEEEabrv,compact}

\end{document}